\documentclass{article}

\newif\ifarxiv
\arxivtrue
\newif\ifslow
\slowfalse

\usepackage{url}

\usepackage{graphicx} 
\graphicspath{{./}}

\usepackage[utf8]{inputenc}

\usepackage{natbib}

\usepackage{amsmath}
\usepackage{bm}

\def\sa{\vspace*{0mm}}
\def\sb{\vspace*{0mm}}
\def\sc{\vspace*{0mm}}

\title{STDP as presynaptic activity times rate of change of postsynaptic activity}

\author{Yoshua Bengio$^1$, Thomas Mesnard, Asja Fischer, \\
  Saizheng Zhang, and Yuhuai Wu\\ 
  Montreal Institute for Learning Algorithms, University of Montreal,\\
  Montreal, QC, H3C 3J7\\
$^1$CIFAR Senior Fellow}

\date{}

\begin{document} 


%

\maketitle

\sb
\begin{abstract}
  We introduce a weight update formula that is expressed only in terms of firing rates and their derivatives and that results in changes consistent with those associated with spike-timing dependent plasticity (STDP) rules and biological observations, even though the explicit timing of spikes is not needed. The new rule changes a synaptic weight in proportion to the product of the presynaptic firing rate and the temporal rate of change of activity on the postsynaptic side. These quantities are interesting for studying theoretical explanation for synaptic changes from a machine learning perspective. In particular, if neural dynamics moved neural activity towards reducing some objective function, then this STDP rule would correspond to stochastic gradient descent on that objective function.
\end{abstract}

\sb
\section{Introduction}
\sb

Biological experiments~\citep{Markram+Sakmann-1995,Gerstner-et-al-1996}
suggest that synaptic change in several types of neurons
and stimulation contexts depends on the relative timing of
presynaptic and postsynaptic spikes.
Learning rules that mimic this observation
are called Spike-Timing Dependent Plasticity (STDP) rules.
In this paper, we investigate a learning rule that
only depends on firing rates and their temporal derivative while
being consistent with the STDP observations, according to
a series of simulations.

This learning rule is motivated by the desire to make it easier
to study machine learning interpretations for STDP, since
artificial neural networks are generally cast in terms of
firing rates. Deep neural networks~\citep{Goodfellow-et-al-2015-Book}
have been extremely successful in several major artificial intelligence
applications in areas such as computer vision, speech recognition,
natural language processing, game playing and robotics. All of these
breakthroughs have been achieved thanks to at least two
ingredients: (a) a deep enough neural network (with enough
layers) and (b) a form of stochastic gradient descent on
an objective function of interest, where the gradient
is obtained by back-propagation~\citep{Rumelhart86b}.

To illustrate the potential use of the proposed STDP rule
in bridging the gap between machine learning based back-propagation
and biology, we show that this STDP rule would correspond
to stochastic gradient descent on an objective function if
the neural activity would move towards reducing that objective function.

As usual, we assume the existence of a non-linear
transformation $\rho$ that is monotonically increasing, from the integrated
activity (the expected membrane potential, averaged over the random effects
of both pre- and postsynaptic spikes) to the actual firing rate.
In deriving a link between STDP and their rate-based learning rule
\citet{Xie+Seung-NIPS1999} started by assuming a particular pattern
relating spike timing and weight change, and then showed that the resulting
weight change could be approximated by the product of presynaptic firing
rate and temporal rate of postsynaptic firing rate. Instead, we go in the
other direction, showing that if the weight change is proportional to the
product of presynaptic firing rate (or simply the presence of a spike) and
the postsynaptic activity, then we recover a
relationship between spike timing and weight change that has the precise
characteristics of the one observed experimentally by biologists. We
present both an easy to understand theoretical justification for this
result, as well as, simulations that confirm it.

\begin{figure}[htpb]
\hspace*{-1mm} \includegraphics[width=0.4\textwidth]{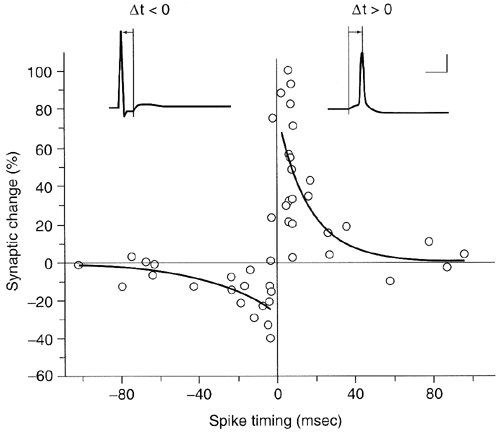}
\hspace*{-1mm}
\includegraphics[width=0.29\textwidth]{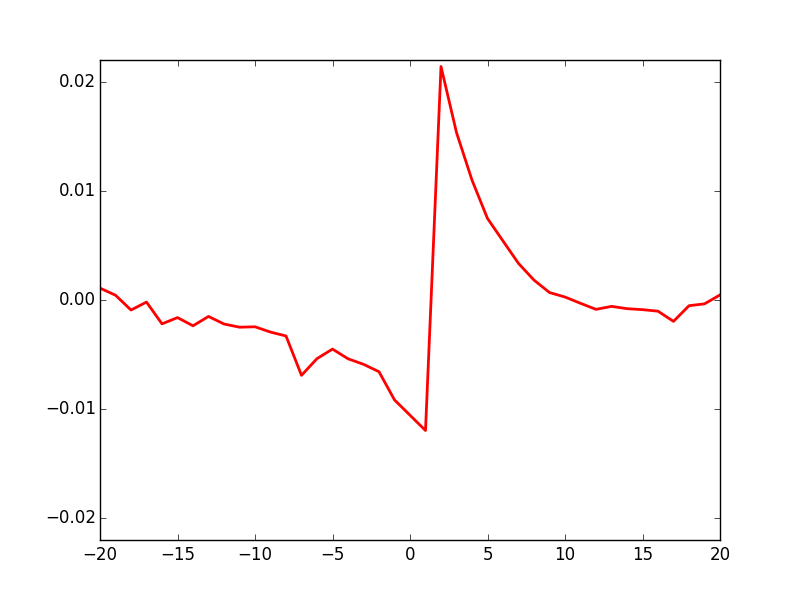} \hspace*{-1.5mm} %
\includegraphics[width=0.29\textwidth]{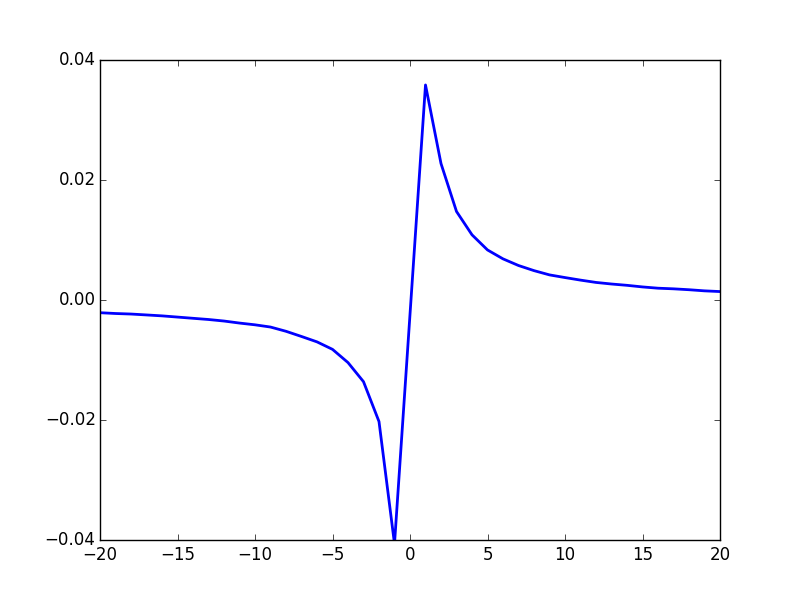}
\caption
{Left:
Biological observation of STDP weight change, vertical axis, for different spike timing
differences (post minus pre), horizontal axis. From~\citet{Shepherd-book2003}, with data from~\citet{Bi+Poo-2001}.
Compare with the result of the simulations using the objective function proposed here (middle).\newline
Middle and right:Spike-based simulation shows that when weight updates follow SGD on the proposed predictive
objective function, we recover the biologically observed relationship between spike timing difference (horizontal axis,
postsynaptic spike time minus presynaptic spike time) and the weight update (vertical axis). Middle: the
weight updates are obtained with the proposed update rule (Eq.~\ref{eq:delta-w-stdp}). Right: the weight
updates are obtained using the nearest neighbor STDP rule.
Compare with the biological finding, left.}
\label{fig:STDP-biology}
\end{figure}

\sb
\section{Spike-timing dependent plasticity}
\sb

Spike-timing dependent plasticity (STDP) is a central subject of research in synaptic plasticity
but much more research is needed to solidify the links between STDP and
a machine learning interpretation of it at the scale of a whole network,
i.e., with ``hidden layers'' which need to receive a useful training signal.
See~\citet{Markram-et-al-2012} for a recent review of the neuroscience
research on STDP. 

We present here a simple view of STDP that has been observed at least in some common
neuron types and preparations. There is a weight change if there is a
presynaptic spike in the temporal vicinity of a postsynaptic spike: 
that change is positive if the postsynaptic
spike happens just after the presynaptic spike (and larger if the timing
difference is small), negative if it happens
just before (and again, larger if the timing difference is small, on the
order of a few milliseconds), as illustrated with the biological
data shown in Figure~\ref{fig:STDP-biology} (left) from~\citet{Bi+Poo-2001}. 
The amount of change decays to zero as the
temporal difference between the two spikes increases beyond a few
tens of milliseconds.
We are thus interested in this temporal window around a presynaptic spike during which
a postsynaptic neuron spikes, before or after the presynaptic spike, 
and this induces a change of the weight.

Keep in mind that the above pattern of spike-timing dependence
is only one aspect of synaptic plasticity. See~\citet{Feldman-2012}
for an overview of aspects of synaptic plastiticy
that go beyond spike timing, including firing rate (as one would expect
from this work) but also synaptic cooperativity (nearby synapses on the
same dendritic subtree) and depolarization (due to multiple consecutive
pairings or spatial integration across nearby locations on the dendrite,
as well as the effect of the synapse's distance to the soma).

\sb
\subsection{Rate-based aspect of STDP}
\sb

Let $s_i$ represent an abstract variable characterizing the integrated
activity of neuron $i$, with $\dot{s}_i$ being its temporal rate of change.
To give a precise meaning to $s_i$, we define
 $\rho(s_i)$ as being the firing rate of neuron $i$, where 
$\rho$ is a bounded non-linear activation function that converts the integrated
voltage potential into a probability of firing.

A hypothesis inspired by~\citet{Xie+Seung-NIPS1999} and~\citet{Hinton-DL2007}
is that the STDP weight change can be associated with the temporal
rate of change of postsynaptic activity, as a proxy for its association
with post minus presynaptic spike times. Both of the above contributions
focus on the rate aspect of neural activity, and this is also the
case of this paper.

Using the notation introduced above,
the proposed equation for the average weight change for the synapse
associated with presynaptic neuron $i$ and postsynaptic neuron $j$
is the following:
\begin{equation}
\sa
 \Delta W_{i,j} \propto \dot{s}_i \rho(s_j) \enspace.
\label{eq:delta-w-stdp}
\sa
\end{equation}

Since stochastic gradient only cares about the {\em average change}, we could as well
have written
\begin{equation}
\sa
\label{eq:delta-w-stdp-spiking}
\Delta W_{i,j} \propto \dot{s}_i \xi_j \enspace,
\sa
\end{equation}
where $\xi_j$ is the binary indicator of a spike from neuron $j$. This works
if we assume that the input spikes are randomly drawn from a Poisson
process with a rate proportional to $\rho(s_j)$, i.e., we ignore 
spike synchrony effects (which we will do in the rest of this paper,
and leave for future work to investigate). Note that in our simulations,
we approximated the Poisson process by performing a discrete-time
simulation with a binomial draw of the binary decision spike versus no spike within the time
interval of interest.

\sc
\begin{figure}[htpb]
  \hspace*{-7mm}\includegraphics[width=0.49\textwidth]{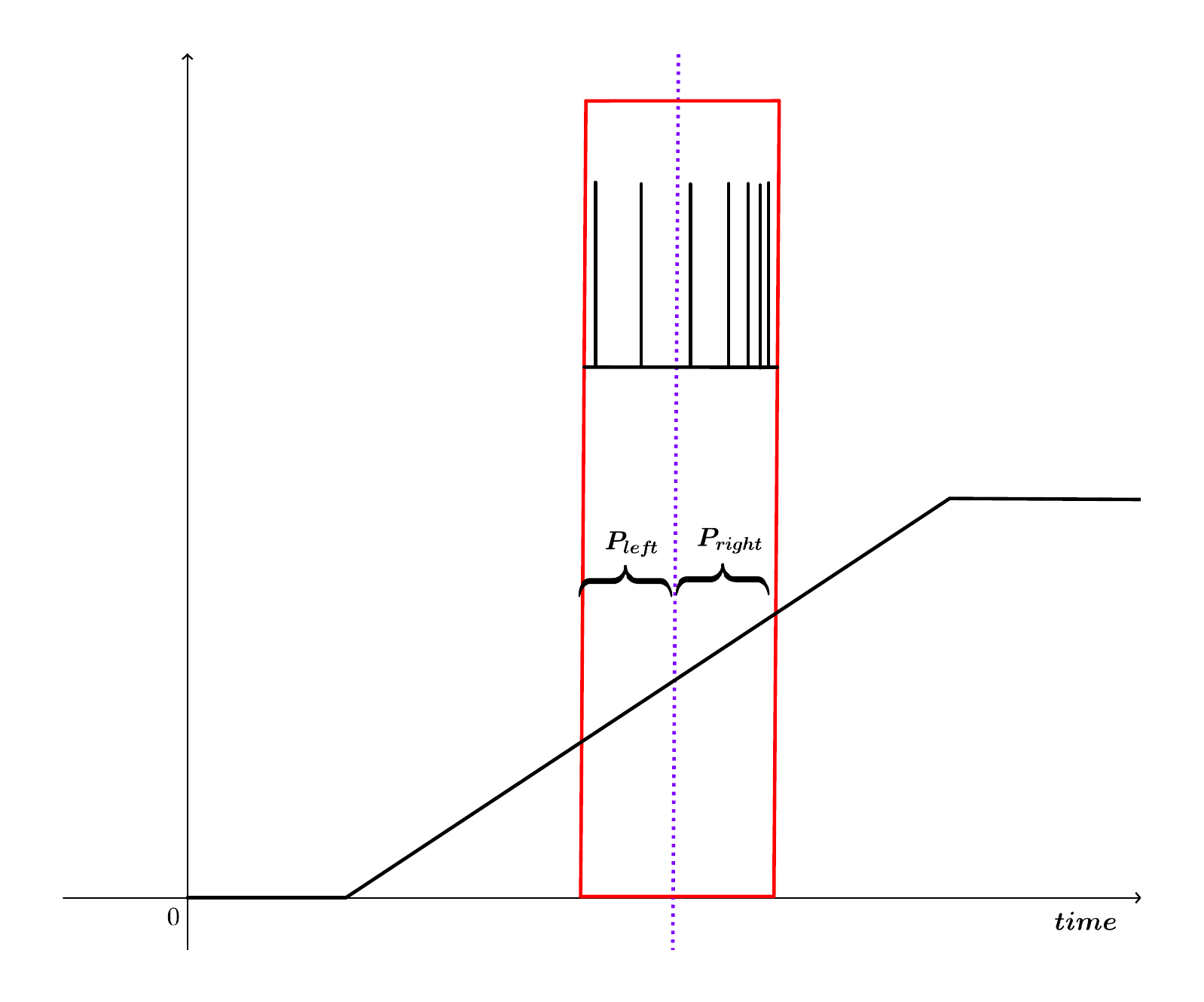}
  \includegraphics[width=.49\textwidth]{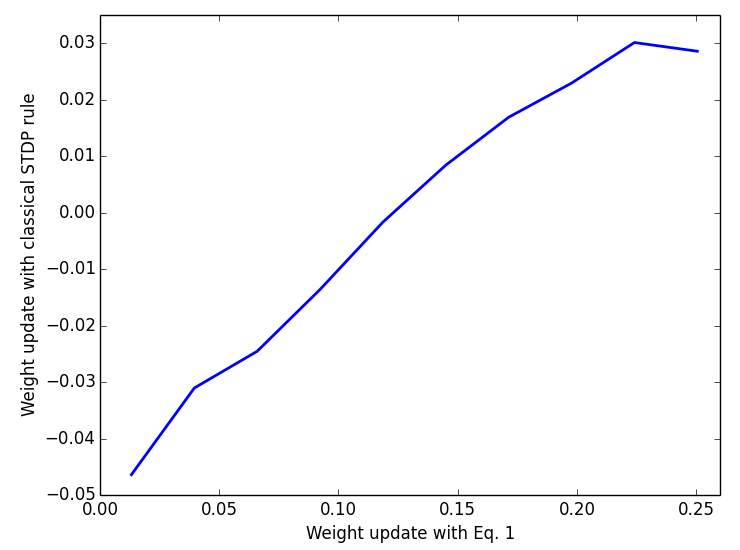}
\vspace*{2mm}
\caption{
Left: When the postsynaptic rate (y-axis) is rising in time (x-axis), 
consider a presynaptic spike (middle
dotted vertical line) and a window of sensitivity before and after (bold red window).
Because the firing probability is greater on the right sub-window, one is more likely
to find a spike there than in the left sub-window, and it is more likely to be close
to the presynaptic spike if that slope is higher, given that when spikes occur
on both sides, no weight update occurs. This induces the appropriate correlation
between spike timing and the temporal slope of the postsynaptic activity level,
which is confirmed by the simulation results of Section~\ref{sec:simulation}
and the results on the right.\newline
Right: The average STDP update according to the STDP nearest neighbor rule
(vertical axis) versus the weight update according to the proposed rule (Eq.~\ref{eq:delta-w-stdp})
(horizontal axis). We see that in average over samples (by binning values of the
x-axis values), the two rules agree closely, in agreement with the visual inspection of
Fig.~\ref{fig:STDP-biology}.
}
\label{fig:stdp}
\end{figure}

\sa
\subsection{Why it works}
\sa

Consider a rising postsynaptic activity level, i.e., $\dot{s}_i>0$,
as in Figure~\ref{fig:stdp} (left)
and a presynaptic spike occurring somewhere during that rise. 
We assume a Poisson process for the postsynaptic spike as well, with a rate
proportional to $\rho(s_i)$.  
According to this assumption, postsynaptic spikes are more likely in
a fixed time window following the presynaptic spike than in a window
of the same length preceding it.
Therefore, if only one spike happens over
the window, it is more likely to be after the presynaptic spike,
yielding a positive spike timing difference, at the same time as a positive
weight change. The situation would be symmetrically reversed if the
activity level was decreasing, i.e., $\dot{s}_i < 0$ and negative spike times
are more likely.

Furthermore, the stronger the slope $\dot{s}_i$, the
greater will this effect be, also reducing the relative spike timing
between the presynaptic spike and stochastically
occurring postsynaptic spikes, possibly just after
or just before the presynaptic spike. This assumes
that when spikes occur on both sides and with about the same time difference,
the effects on $W_{i,j}$ cancel each
other. Of course, the above reasoning runs directly in reverse when the
slope of $s_i$ is negative, and one gets negative updates, in average. To
validate these hypothesis, we ran the simulations presented in
Section~\ref{sec:simulation}.  They confirm these hypotheses and show that
Eq.~\ref{eq:delta-w-stdp} yields a relationship between spike timing
difference and weight change that is consistent with biological
observations (Fig.~\ref{fig:STDP-biology}).

\sb
\section{Simulation results}
\label{sec:simulation}
\sb

\subsection{Method}
\sa

We simulate random variations in a presynaptic neural firing rate $\rho(s_j)$ as well as 
random variations
in the postsynaptic firing rates $\rho(s_i)$ induced by an externally driven voltage. By
exploring many configurations of variations and levels at pre- and postsynaptic sides, we
hope to cover the possible natural variations. We generate and record
pre- and postsynaptic spikes sampled according to a binomial at each discrete $t$ with probability proportional
to $\rho(s_i)$ and $\rho(s_j)$ respectively, and record $\dot{s}_i$ as well, in order to implement
either a classical nearest neighbor STDP update rule or Eq.~\ref{eq:delta-w-stdp},
\footnote{
Python scripts for those simulations are available at
\linebreak[4]{\small \tt http://www.iro.umontreal.ca/$\tilde{~}$bengioy/src/STDP$\_$simulations}.}
 The nearest neighbor STDP rule 
is as follows. For every presynaptic spike, we consider a window of 20 time steps
before and after. If there is one or more postsynaptic spike in both left and right
windows, or no postsynaptic spike at all, the weight is not updated. Otherwise, we
measure the time difference between the closest postsynaptic spike (nearest neighbor)
and the presynaptic spike and compute the weight change using the current variable values.
If both spikes coincide, we make no weight change. 
To compute the appropriate averages, 500 random sequences
of rates are generated, each of length 160 time steps, and 1000 randomly sampled spike trains
are generated according to these rates.

For measuring the effect of weight changes, we measure the average squared rate of
change $E[||\dot{s}_i^2||]$ in two conditions: with weight changes (according
to Eq.~\ref{eq:delta-w-stdp}), and without.

\sa
\subsection{Results}
\sa

\begin{figure}[htpb]
\centerline{\includegraphics[width=0.7\textwidth]{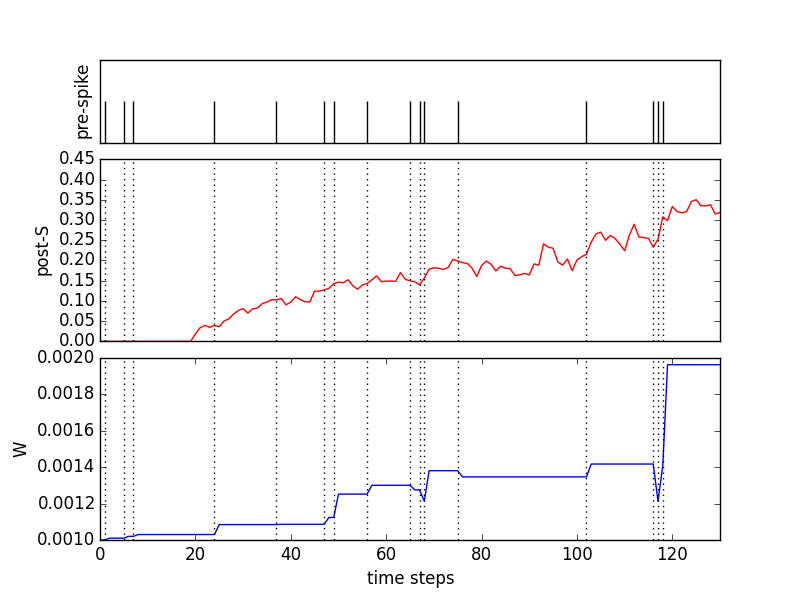}}
\caption{Example of rate and spike sequence generated in the simulations, along with
weight changes according to the spike-dependent variant of our update rule, Eq.~\ref{eq:delta-w-stdp-spiking}.
Top: presynaptic spikes $\xi_j$ (which is when a weight change can occur).
Middle: integrated postsynaptic activity $s_j$.
Bottom: value of the updated weight $W_{i,j}$.
}
\label{fig:examples-spikes-sequences}
\end{figure}

Examples of the spike sequences and underlying pre- and postsynaptic states $s_i$
and $s_j$ are illustrated in Fig.~\ref{fig:examples-spikes-sequences}.

Fig.~\ref{fig:STDP-biology} (middle and right) shows the results of these simulations, comparing
the weight change obtained at various spike timing differences for Eq.~\ref{eq:delta-w-stdp}
and for the nearest neighbor STDP rule, both matching well the biological data
(Fig.~\ref{fig:STDP-biology}, left). Fig.~\ref{fig:stdp} shows that
both update rules are strongly correlated, in the sense that for a given amount
of weight change induced by one, we observe in average a linearly proportional weight
change by the other.

\sa
\subsection{Link to Stochastic Gradient Descent and Back-Propagation}
\sa

Let us consider the common simplification in which
postsynaptic neural activity $s_i$ is obtained as a sum with the usual terms
proportional to the product of synaptic weight ($W_{i,j}$) and presynaptic
firing rate ($\rho(s_j(t-1)$), i.e., in discrete time,
\[
  s_i(t)=\alpha+\beta \sum_j W_{i,j} \rho(s_j(t-1)).
  \]
  from some quantities $\alpha$ and $\beta$. In that case,
  \[
  \frac{\partial s_i(t)}{\partial W_{i,j}} = \beta \rho(s_j(t-1)).
  \]
  Stochastic gradient descent on $W$ with respect to an objective function $J$
  would follow
  \begin{align}
    \Delta W_{i,j} \propto \frac{\partial J}{\partial W_{i,j}} &= \frac{\partial J}{\partial s_i(t)} \frac{\partial s_i(t)}{\partial W_{i,j}} \nonumber \\
    &= \frac{\partial J}{\partial s_i(t)} \rho(s_j(t-1)).
  \end{align}
  Hence, if
  \begin{equation}
    \label{eq:neurons-move-to-reduce-error}
  \frac{\partial J}{\partial s_i(t)} \propto \dot{s}_i(t),
  \end{equation}
  our STDP rule (Eq.~\ref{eq:delta-w-stdp}) would produce
  stochastic gradient on $J$.

  The assumption in Eq.~\ref{eq:neurons-move-to-reduce-error} and the above consequence
  was first suggested by~\citet{Hinton-DL2007}:
  neurons would change their average firing rate so as make the network
  as a whole produce configurations corresponding to better values
  of our objective function. This would make STDP do gradient descent on
  the prediction errors made by the network.

  But how could neural dynamics have that property? That question is
  not completely answered, but as shown in~\citet{Bengio-arxiv2015}, a network with symmetric feedback
  weights would have the property that a small perturbation of ``output units'' towards better
  predictions would propagate to internal layers such that hidden
  units $s_i$ would move to approximately follow the gradient of the prediction error $J$
  with respect to $s_i$, making our STDP rule correspond
  approximately to gradient descent on the prediction error.
  The experiments reported by ~\citet{Scellier+Bengio-arxiv2016} have actuallly shown
  that these approximations work and enable a supervised multi-layer neural network to be trained.
  
\sb
\section{Related work}
\sb

The closest work to this paper is probably that of~\citet{Xie+Seung-NIPS1999}, which we have
already discussed. Notable additions to this work include demonstrating
that the spike-timing to weight change relationship is a {\em consequence}
of Eq.~\ref{eq:delta-w-stdp}, rather than the other way around (and a small
difference in the use of $ds/dt$ versus $d\rho(s)/dt$).

There are of course many other papers on theoretical interpretations of STDP,
and the reader can find many references in~\citet{Markram-et-al-2012}, but more 
work is needed to explore the connection of STDP to machine learning.
Many approaches~\citep{Fiete+Seung-2006,Rezende+Gerstner-2014} that propose how
hidden layers of biological neurons could get credit assignment rely on variants of the REINFORCE 
algorithm~\citep{Williams-1992}
to estimate the gradient of a global objective function (basically by correlating
stochastic variations at each neuron with the changes in the global objective).
Although this principle is simple, it is not clear that it will scale to very
large networks due to the linear growth of the variance of the estimator with the number
of neurons. It is therefore tempting to explore other avenues.

The proposed rule can be contrasted with theoretical synaptic learning rules
which tend to be based on the Hebbian product of pre- and postsynaptic activity,
such as the BCM rule~\citep{Bienenstock82,Intrator+Cooper-1992}.
The main difference is that the rule studied
here involves the temporal derivative of the postsynaptic activity, rather than
the actual level of postsynaptic activity.

\section{Conclusion and Open Questions}

We have shown through simulations and a qualitative argument that updating
weights in proportion to 
the rate of change of postsynaptic activity times the presynaptic
activity yielded behavior similar to the STDP observations, in terms
of spike timing differences.

This is consistent with a view introduced by~\citet{Hinton-DL2007}
that neurons move towards minimizing some prediction error because
it would make this STDP rule perform stochastic gradient descent on that
prediction error. In parallel to this work, ~\citet{Bengio-arxiv2015} has
shown how early propagation of perturbations in a recurrent network
behaves like gradient propagation in a layered network, and could
compute the gradients necessary for learning using our proposed
STDP update. ~\citet{Scellier+Bengio-arxiv2016} have actuallly shown
through experiments using this approach that it was possible to
train deep supervised neural networks.

Many open questions also remain on the side of biological plausibility.
STDP only describes one view of synaptic plasticity, and it may be different
in different neurons and preparation.
A particularly interesting type of result to consider in the future
regards the relationship between more than two
spikes induced statistically by the proposed STDP rule.
It would be interested to compare the statistics of triplets
or quadruplets of spikes timings and weight change under the proposed rule.
against the corresponding biological observations~\citep{Froemke+Dan-2002}.

Future work should also attempt to reconcile the theory with
neuroscience experiments showing more complex relationships between spikes
and weight change, involving factors other than timing, such as firing rate,
of course, but also synaptic cooperativity and depolarization~\citep{Feldman-2012}.
In addition to these factors, some synaptic changes do not conform to
the STDP pattern, either in sign (anti-Hebbian STDP) or in nature
(for example incorporating a general Hebbian element as in the BCM rule).
In this regard, although this work suggests that some
spike timing behavior may be explained purely in terms of the rates trajectory,
it remains an open question if (and where) precise timing of spikes remains
essential to explain network-level learning behavior from a machine learning
perspective. It is interesting to note in this regard how the BCM rule
was adapted to account for the relative timing of spike triplets~\citep{Gjorgjievaa-et-al-2011}.

\ifarxiv
\section*{Acknowledgments} 
 
The authors would like to thank Benjamin Scellier, Dong-Hyun Lee, Jyri Kivinen, Jorg Bornschein, Roland Memisevic and Tim Lillicrap 
for feedback and discussions, as well as NSERC, CIFAR, Samsung
and Canada Research Chairs for funding, as well as Compute Canada
for computing resources.
\else
\fi

\bibliography{strings,ml}

\bibliographystyle{natbib}

\end{document}